\documentclass{tlp}

\usepackage{amsmath}
\usepackage{graphicx}
\usepackage{multirow}
\usepackage[disable]{todonotes} 

\usepackage{makecell}

\usepackage[frozencache,cachedir=.]{minted}
\setminted{ linenos,  numberblanklines=false,  firstnumber=last,  escapeinside=|| }

\usepackage{subcaption}

\PassOptionsToPackage{rgb}{xcolor}
\usepackage{tikz}
\usetikzlibrary{arrows.meta,patterns}
\usetikzlibrary{ipe}

\newtheorem{definition}{Definition}

\newcommand{\absorbing}{controlling}
\newcommand{\Absorbing}{Controlling}
\newcommand{\neutral}{non-controlling}

\newcommand{\NeutralVar}{NonControlling} \newcommand{\neutralcode}{non\_controlling}

\usepackage{acronym}
\newacro{ASP}{Answer Set Programming}
\newacro{CLP(FD)}{Constraint Logic Programming on Finite Domains}
\newacro{MILP}{Mixed-Integer Linear Programming}
\newacro{IC}{Integrity Constraint}
\newacro{PI}{Primary Input}
\newacro{PO}{Primary Output}
\newacro{STA}{Static Timing Analysis}
\newacro{INCV}{Input Non Controlling Value}
\newacro{ICV}{Input Controlling Value}
\newacro{OCV}{Output Controlled Value} \newacro{ONCV}{Output Non Controlled Value} \newacro{LST}{Latest Stabilization Time}
\newacro{EAT}{Early Arrival Time}
\newacro{ALU}{Arithmetic Logic Unit}

\newcommand{\vettore}[1]{{\ensuremath{{\mathbf{V}_#1}}}}  \newcommand{\Ninput}{{\ensuremath{n_i}}} \newcommand{\Noutput}{{\ensuremath{n_o}}} \newcommand{\inputs}{{\ensuremath{{\mathbf{I}}}}} \newcommand{\outputs}{{\ensuremath{{\mathbf{O}}}}} \newcommand{\val}[2]{{\ensuremath{v^#1_#2}}} \usepackage{ifthen}
\newcommand{\tempo}[2]{{\ensuremath{\ifthenelse{\equal{#2}{1}}{e^#1}{l^#1}}}}

\newcommand{\tempos}[2]{{\ensuremath{\ifthenelse{\equal{#2}{1}}{e^#1_*}{l^#1_*}}}}

\newcommand{\segnale}[1]{{\ensuremath{#1}}}

\begin{document}

\lefttitle{Bertagnon, Dalpasso, Favalli and Gavanelli}

\jnlPage{1}{8}
\jnlDoiYr{2021}
\doival{10.1017/xxxxx}

\title[Fine-grained Timing Analysis of Digital ICs in ASP]{Fine-grained Timing Analysis of Digital Integrated Circuits in Answer Set Programming
}

\begin{authgrp}

\author{\sn{Bertagnon} \gn{Alessandro}}
\affiliation{Department of Environmental and Prevention Sciences, University of Ferrara}
\author{\sn{Dalpasso} \gn{Marcello}}
\affiliation{DEI - University of Padova}
\author{\sn{Favalli} \gn{Michele}}
\author{\sn{Gavanelli} \gn{Marco}}
\affiliation{Department of Engineering, University of Ferrara}

\end{authgrp}

\history{\sub{18 04 2025;} \rev{22 06 2025;} \acc{08 07 2025}}

\maketitle

\begin{abstract}
In the design of integrated circuits, one critical metric is the maximum delay introduced by combinational modules within the circuit.
This delay is crucial because it represents the time required to perform a computation:
in an Arithmetic-Logic Unit it represents the maximum time taken by the circuit to perform an
arithmetic operation. When such a circuit is part of a larger, synchronous system, like a CPU, the maximum
delay directly impacts the maximum clock frequency of the entire system.
Typically, hardware designers use Static Timing Analysis to compute an upper bound of the maximum delay because it can
be determined in polynomial time. However, relying on this upper bound can lead to suboptimal
processor speeds, thereby missing performance opportunities.
In this work, we tackle the challenging task of computing the actual maximum delay, rather
than an approximate value. Since the problem is computationally hard, we model it in Answer
Set Programming (ASP), a logic language featuring extremely efficient solvers. We propose  non-trivial encodings of the problem into ASP.
Experimental results show that ASP is a viable solution to address complex problems in hardware design.

\end{abstract}


\section{Introduction}
\label{sec:introduction}
\definecolor{turquoise}{rgb}{0.251,0.878,0.816}

In the design of integrated circuits, one critical metric is the maximum delay introduced by combinational modules within the circuit \citep{sta0,sta1,k1994,falsepath,pathselection,sta,DBLP:conf/lpnmr/AndresSGSBS13,8299562,10457948}.
For example, in digital synchronous circuits  the maximum delay in the combinational (i.e. acyclic)
logic blocks is necessary to determine the clock period. 
For instance, an \ac{ALU} is a crucial component of a CPU; it is a combinational circuit (its output depends only on the inputs, 
and it does not have an internal memory) but its speed influences the overall 
speed of the CPU, which is a synchronous sequential circuit.
When designing an \ac{ALU}, the designer 
must determine the time required to produce
the final output after a new input is provided to the circuit;
the maximum possible delay influences the maximum clock rate the CPU can run at.
This step is first performed at early design steps using simplified delay models 
for gates in order to drive possible circuit optimizations, and finally it is performed after the physical design using more complex delay models that use accurate electrical level information. 
For simplicity, we will consider the basic case, although the proposed approach can be extended to more accurate delay models.

An approximate approach to maximum delay computation 
is given by \ac{STA} \citep{sta0,sta1,sta}, which computes this delay as the longest 
path in a directed acyclic graph. The circuit is  traversed from 
primary inputs (PIs) to primary outputs (POs); the static arrival time for the output of each gate is 
the sum of 
the maximum 
arrival time of its inputs plus the gate propagation delay. 
This method is static because it does not consider the actual logic 
values within the circuit. Therefore, it provides possibly pessimistic 
results since there might exist no input configuration that propagates 
transitions through the computed longest paths. 

\begin{figure}
\centerline{\ \includegraphics[width=0.9\textwidth]{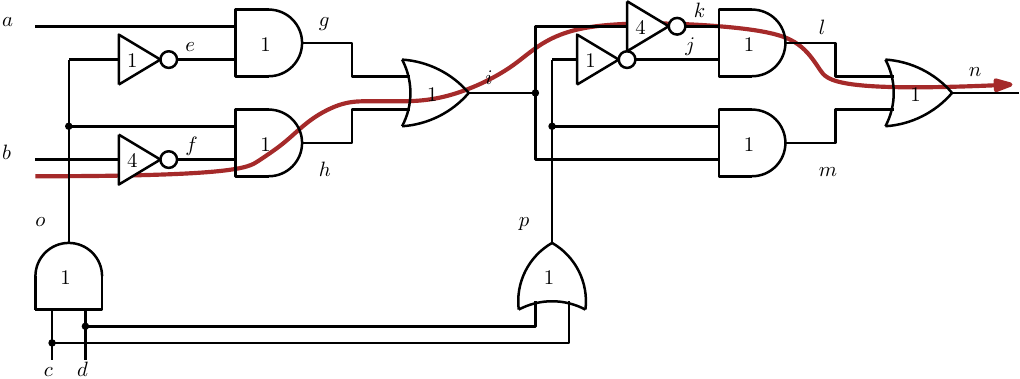}}
\caption[x]{\protect Example of circuit featuring a false path that leads to pessimism in \ac{STA}. }
\label{staticta}
\end{figure}

As an example, consider the circuit in Fig.~\ref{staticta}, 
where the delay introduced by each gate is represented by a number inside it (in arbitrary time units). 
In such a circuit, the maximum delay computed by \ac{STA} is 12, with  
the path 
$b\textrm{-} f\textrm{-} h\textrm{-} i\textrm{-} k\textrm{-} l\textrm{-} n$.
Paths featuring the maximum delay are called critical paths \citep{k1994}. 
However, no sequence 
of input vectors exist that make a transition propagate through such a path. 
In fact, if signal $o$ is 0 (false) then $h$ is always 0, independently of $b$,
while if $p$ is 1 (true) the output $l$ is always 0, independently from the path highlighted in red.
So, the only possibility for a signal from $b$ to influence the output $n$ would be that $o=1$ and $p=0$,
which is impossible since $o=c \land d$ and $p=c \lor d$.
The maximum delay of such a circuit, which
can be computed with more accurate approaches and by means of the proposed method, is 10.

To refine such results with the aim to meet the aggressive timing 
requirements of today's circuits, several approaches consider circuit paths 
and try to sensitize them  to prove that transitions can propagate 
through them. If this operation is possible, the path is said to be true, 
otherwise, the path is said to be false (such as the path indicated in Fig.~\ref{staticta}) because it cannot 
contribute to the maximum circuit delay \citep{falsepath}.  
For computational reasons, only the longest paths are typically 
considered \citep{pathselection}.
The way in which a path is sensitized determines the quality of the      
results achievable by such timing verification techniques.

For example, \cite{DBLP:conf/lpnmr/AndresSGSBS13} use  \ac{ASP} to compute the maximum delay in a circuit.
They search for a sequence of two input vectors 
such that when the circuit passes from the first bit vector to the second,
a transition is propagated through the longest sensitizable path.
A path is sensitizable if all the gates in the path flip (change their output from 1 to 0 or vice-versa)
when the input changes from the first input vector to the second.

\begin{figure}
\centering
\begin{subfigure}{0.65\textwidth}
    \includegraphics[width=\textwidth]{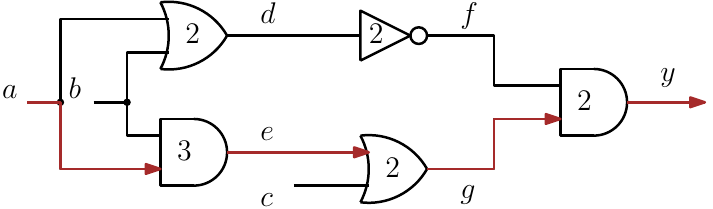}
    \vspace{0.78cm}
    \caption{}
    \label{ex}
\end{subfigure}\hfill
\begin{subfigure}{0.25\textwidth}
    \includegraphics[width=\textwidth]{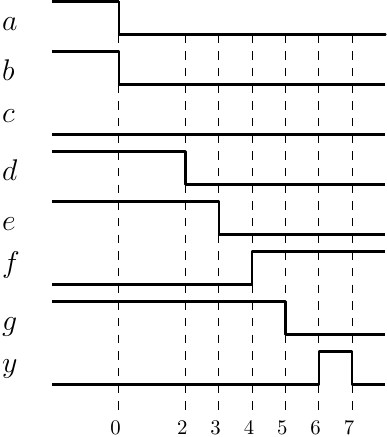}
    \caption{}
    \label{fig:bench0_wave}
\end{subfigure}
\caption{(a) Example circuit and (b) corresponding signal waveforms.}
\end{figure}

Consider, for instance, the circuit in Fig.~\ref{ex}.
The path $a\textrm{-}d\textrm{-}f\textrm{-}y$ can be sensitized, in fact if we take as first input vector
$[a=0,b=0,c=1]$ and as second
$[a=1,b=1,c=0]$, all the gates in the path
switch their output value ($d$, $f$ and $y$ change value).
The total delay on this path is the sum of the delays in the path: $2+2+2=6$.

Instead, the path $a\textrm{-}e\textrm{-}g\textrm{-}y$ cannot be statically sensitized, in fact in order for the input 
$a$ to change the output $e$ of the first AND gate, it is required that
$b=1$, and for the same reason $c=0$ and $f=1$. It is easy to find that $b=1 \rightarrow f=0$, thus preventing a transition on $a$ to reach the output. Such path has delay=7, and cannot be found with  path sensitization.
Unfortunately, although the longest sensitizable path has length 6,
this does not mean that after 6 time units the final value has been correctly computed, or that
there cannot be unpleasant effects happening after such time.

Consider, for example, the switch from $[a=1,b=1,c=0]$ to $[a=0,b=0,c=0]$; the signals' waveforms are shown in Figure~\ref{fig:bench0_wave}. At time 0, $a$ and $b$ switch from level 1 to 0;
2 time units later, $d$ flips to 0, and at time 4 signal $f$ goes to 1.
Signal $e$ changes at time 3, causing $g$ to flip to 0 at time 5.
Note that between time 4 and 5, both $f$ and $g$ have value 1: this makes the final AND gate switch its output
from 0 to 1 in instant 6. This is not the final output value: in fact at time 5 the signal $g$ goes to 0, making the $out$ signal go to its final level only at time 7.
The temporary switching of a signal is called in the literature a {\em hazard};
the circuit in Figure~\ref{ex} has a {\em $1$-sized hazard} of the $y$  signal, since it lasts 1 time unit.

In the next section we propose a model of a circuit that also takes into consideration hazards when computing the maximum delay.

 \section{Problem Description}
\label{sec:problem}

Unlike path based approaches, which constitute a large majority 
in the field of timing 
verification to improve \ac{STA} results, our approach does not explicitly consider 
paths. Instead, it directly models the  dynamic conditions 
occurring within the circuit for the possible input configurations.

Let us consider a combinational circuit with a set \inputs\ of input signals, where the number of inputs is denoted by $\Ninput = |\inputs|$, and a set \outputs\ of output signals, with
$\Noutput = |\outputs|$.

To compute the maximum possible delay of the circuit, we study its behavior between two arbitrary input vectors, $\vettore{1}$ and $\vettore{2} \in \{0,1\}^\Ninput$, which are applied one after the other to the input signals.

The state of each signal $s$ in the circuit is represented by a 4-tuple $\langle \val{s}{1}, \val{s}{2}, \tempo{s}{1}, \tempo{s}{2} \rangle$, where
1) $\val{s}{1} \in \{0,1\}$ is the logic value to which $s$ stabilizes when the first input vector is applied; 2) $\val{s}{2} \in \{0,1\}$ denotes the value to which $s$ eventually stabilizes after applying the second input vector; 
3) $\tempo{s}{1} \in [-1,T]$ denotes the  earliest time at which a transition away from $\val{s}{1}$ may occur; 
4) $\tempo{s}{2} \in [-1,T]$ denotes the latest time by which $s$ stabilizes to $\val{s}{2}$.
$T$ denotes a value much larger than the expected maximal delay of the considered combinational circuit.

The intuitive meaning is that initially the first input vector is applied, and the signal $s$ stabilizes to 
the logical value $\val{s}{1}$. When the second input vector is introduced, the signal  does not change immediately, due to   propagation delays  in the circuit. It remains at $\val{s}{1}$ until the early arrival time \tempo{s}{1}, after which it might have spurious variations. Eventually, by time  \tempo{s}{2} (called the latest stabilization time), it becomes stable to the final value $\val{s}{2}$, which depends only on the second input vector.

The objective is to determine the worst-case delay, defined as the maximum stabilization time across all outputs, over all possible pairs of input vectors
$\vettore{1}$ and $\vettore{2}$:
$$\max_{\vettore{1}, \vettore{2} \in \{0,1\}^\Ninput} \{\tempo{o}{2} \mid o \in \outputs\}.$$

The gate output state is computed as a function of gate input states.
Let us consider a NAND gate with output signal $ y$ and $ a, b$ as inputs.
The input/output relationships for the two considered test vectors are described as $\val{y}{i}=\neg(\val{a}{i} \wedge \val{b}{i})$ for each of the two input vectors (represented by the index $i\in\{1,2\}$).

The arrival times of the gate output are a function of the input logic values and of the arrival times of the gate inputs.
Let us instantiate the computation of \tempo{y}{1}
and \tempo{y}{2}
in a gate delay model featuring the propagation delay $d$ as the only parameter; as already said, more complex delay models can be easily accounted for.

To this purpose, we first introduce two auxiliary variables \tempos{y}{1} and \tempos{y}{2}
that intuitively represent the time in which the gate may start commuting (\tempos{y}{1}) and that in which the output becomes stable (\tempos{y}{2}) without considering the delay introduced by the gate itself.
\begin{subequations}
\begin{align}
\val{a}{1}\wedge \val{b}{1}  \ \rightarrow \ & \tempos{{y}}{1}=\min\{\tempo{a}{1},\tempo{b}{1}\} \\
\neg \val{a}{1}\wedge \neg \val{b}{1}  \ \rightarrow \ & \tempos{{y}}{1}=\max\{\tempo{a}{1},\tempo{b}{1}\} \\
\val{a}{1}\wedge \neg \val{b}{1}  \ \rightarrow \ & \tempos{{y}}{1}=\tempo{b}{1}  \\
\neg \val{a}{1}\wedge \val{b}{1}  \ \rightarrow \ & \tempos{{y}}{1}=\tempo{a}{1}  \\
\val{a}{2} \wedge \val{b}{2}   \ \rightarrow \ & \tempos{{y}}{2}=\max\{\tempo{a}{2},\tempo{b}{2}\} \\
\neg \val{a}{2} \wedge \neg \val{b}{2}   \ \rightarrow \ & \tempos{{y}}{2}=\min\{\tempo{a}{2},\tempo{b}{2}\} \\
\neg \val{a}{2} \wedge \val{b}{2}   \ \rightarrow \ & \tempos{{y}}{2}=\tempo{a}{2}  \\
\val{a}{2} \wedge \neg \val{b}{2}   \ \rightarrow \ & \tempos{{y}}{2}=\tempo{b}{2}
\end{align}
\label{eq1}
\end{subequations}

\tikzstyle{ipe stylesheet} = [
  ipe import,
  even odd rule,
  line join=round,
  line cap=butt,
  ipe pen normal/.style={line width=0.4},
  ipe pen heavier/.style={line width=0.8},
  ipe pen fat/.style={line width=1.2},
  ipe pen ultrafat/.style={line width=2},
  ipe pen normal,
  ipe mark normal/.style={ipe mark scale=3},
  ipe mark large/.style={ipe mark scale=5},
  ipe mark small/.style={ipe mark scale=2},
  ipe mark tiny/.style={ipe mark scale=1.1},
  ipe mark normal,
  /pgf/arrow keys/.cd,
  ipe arrow normal/.style={scale=7},
  ipe arrow large/.style={scale=10},
  ipe arrow small/.style={scale=5},
  ipe arrow tiny/.style={scale=3},
  ipe arrow normal,
  /tikz/.cd,
  ipe arrows, <->/.tip = ipe normal,
  ipe dash normal/.style={dash pattern=},
  ipe dash dotted/.style={dash pattern=on 1bp off 3bp},
  ipe dash dashed/.style={dash pattern=on 4bp off 4bp},
  ipe dash dash dotted/.style={dash pattern=on 4bp off 2bp on 1bp off 2bp},
  ipe dash dash dot dotted/.style={dash pattern=on 4bp off 2bp on 1bp off 2bp on 1bp off 2bp},
  ipe dash normal,
  ipe node/.append style={font=\normalsize},
  ipe stretch normal/.style={ipe node stretch=1},
  ipe stretch normal,
  ipe opacity 10/.style={opacity=0.1},
  ipe opacity 30/.style={opacity=0.3},
  ipe opacity 50/.style={opacity=0.5},
  ipe opacity 75/.style={opacity=0.75},
  ipe opacity opaque/.style={opacity=1},
  ipe opacity opaque,
]

\definecolor{red}{rgb}{1,0,0}
\definecolor{blue}{rgb}{0,0,1}
\definecolor{green}{rgb}{0,1,0}
\definecolor{yellow}{rgb}{1,1,0}
\definecolor{orange}{rgb}{1,0.647,0}
\definecolor{gold}{rgb}{1,0.843,0}
\definecolor{purple}{rgb}{0.627,0.125,0.941}
\definecolor{gray}{rgb}{0.745,0.745,0.745}
\definecolor{brown}{rgb}{0.647,0.165,0.165}
\definecolor{navy}{rgb}{0,0,0.502}
\definecolor{pink}{rgb}{1,0.753,0.796}
\definecolor{seagreen}{rgb}{0.18,0.545,0.341}
\definecolor{turquoise}{rgb}{0.251,0.878,0.816}
\definecolor{violet}{rgb}{0.933,0.51,0.933}
\definecolor{darkblue}{rgb}{0,0,0.545}
\definecolor{darkcyan}{rgb}{0,0.545,0.545}
\definecolor{darkgray}{rgb}{0.663,0.663,0.663}
\definecolor{darkgreen}{rgb}{0,0.392,0}
\definecolor{darkmagenta}{rgb}{0.545,0,0.545}
\definecolor{darkorange}{rgb}{1,0.549,0}
\definecolor{darkred}{rgb}{0.545,0,0}
\definecolor{lightblue}{rgb}{0.678,0.847,0.902}
\definecolor{lightcyan}{rgb}{0.878,1,1}
\definecolor{lightgray}{rgb}{0.827,0.827,0.827}
\definecolor{lightgreen}{rgb}{0.565,0.933,0.565}
\definecolor{lightyellow}{rgb}{1,1,0.878}
\definecolor{black}{rgb}{0,0,0}
\definecolor{white}{rgb}{1,1,1}

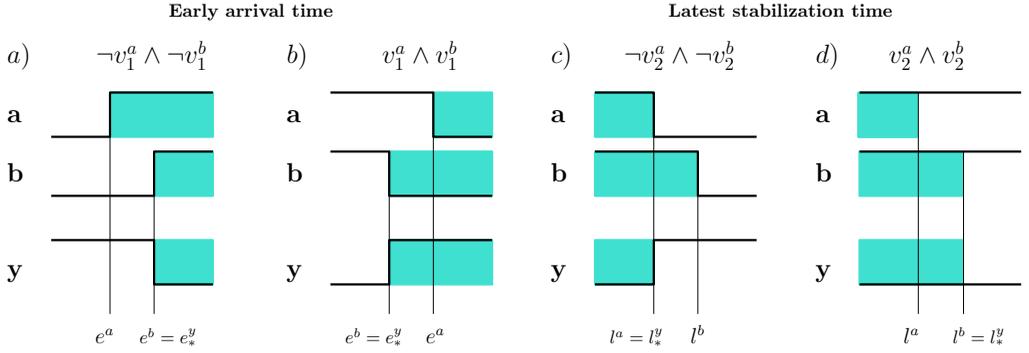
\begin{figure}
\centering
\resizebox{\textwidth}{!}{\begin{tikzpicture}[ipe stylesheet]
\filldraw[turquoise, ipe pen fat]
    (124, 752) rectangle (124, 752);
  \node[ipe node, font=\Large]
     at (24, 760) {$$\bf a$ $};
  \node[ipe node, font=\Large]
     at (24, 728) {$$\bf b$$};
  \node[ipe node, font=\Large]
     at (24, 676) {$$\bf y $ $};
  \node[ipe node, font=\large]
     at (72, 640) {$\tempo{a}{1} $};
  \node[ipe node]
     at (96, 640) {$\tempo{b}{1}=\tempos{{y}}{1} $};
  \filldraw[turquoise, ipe pen fat]
    (276, 752) rectangle (276, 752);
  \node[ipe node, font=\Large]
     at (176, 760) {$$\bf a$ $};
  \node[ipe node, font=\Large]
     at (176, 728) {$$\bf b$$};
  \node[ipe node, font=\Large]
     at (176, 676) {$$\bf y $ $};
  \node[ipe node, font=\large]
     at (252, 640) {$\tempo{a}{1}  $};
  \node[ipe node]
     at (208, 640) {$\tempo{b}{1}=\tempos{{y}}{1} $};
  \filldraw[turquoise, ipe pen fat]
    (420, 752) rectangle (420, 752);
  \node[ipe node, font=\Large]
     at (320, 760) {$$\bf a$ $};
  \node[ipe node, font=\Large]
     at (320, 728) {$$\bf b$$};
  \node[ipe node, font=\Large]
     at (320, 676) {$$\bf y $ $};
  \node[ipe node, font=\large]
     at (396, 640) {$\tempo{b}{2}  $};
  \node[ipe node]
     at (352, 640) {$\tempo{a}{2}=\tempos{{y}}{2} $};
  \filldraw[turquoise, ipe pen fat]
    (80, 776) rectangle (136, 752);
  \filldraw[turquoise, ipe pen fat]
    (104, 744) rectangle (136, 720);
  \filldraw[turquoise, ipe pen fat]
    (104, 696) rectangle (136, 672);
  \filldraw[turquoise, ipe pen fat]
    (256, 776) rectangle (288, 752);
  \filldraw[turquoise, ipe pen fat]
    (232, 744) rectangle (288, 720);
  \filldraw[fill=turquoise]
    (104, 744)
     -- (104, 656);
  \filldraw[turquoise, ipe pen fat]
    (232, 696) rectangle (288, 672);
  \filldraw[fill=turquoise]
    (232, 744)
     -- (232, 656);
  \filldraw[turquoise, ipe pen fat]
    (344, 776) rectangle (376, 752);
  \filldraw[turquoise, ipe pen fat]
    (344, 744) rectangle (400, 720);
  \filldraw[turquoise, ipe pen fat]
    (344, 696) rectangle (376, 672);
  \filldraw[fill=turquoise]
    (400, 744)
     -- (400, 656);
  \filldraw[fill=turquoise]
    (80, 776)
     -- (80, 656);
  \filldraw[fill=turquoise]
    (376, 776)
     -- (376, 656);
  \filldraw[fill=turquoise]
    (256, 752)
     -- (256, 656);
  \filldraw[turquoise, ipe pen fat]
    (564, 752) rectangle (564, 752);
  \node[ipe node, font=\Large]
     at (464, 760) {$$\bf a$ $};
  \node[ipe node, font=\Large]
     at (464, 728) {$$\bf b$$};
  \node[ipe node, font=\Large]
     at (464, 676) {$$\bf y $ $};
  \node[ipe node, font=\large]
     at (512, 640) {$\tempo{a}{2}  $};
  \node[ipe node]
     at (538.714, 640.169) {$\tempo{b}{2}=\tempos{{y}}{2} $};
  \filldraw[turquoise, ipe pen fat]
    (488, 744) rectangle (544, 720);
  \filldraw[turquoise, ipe pen fat]
    (487.307, 775.8765) rectangle (519.307, 751.8765);
  \filldraw[turquoise, ipe pen fat]
    (488, 696) rectangle (544, 672);
  \filldraw[fill=turquoise]
    (544.6173, 744)
     -- (544.6173, 656);
  \node[ipe node, anchor=north west]
     at (112, 824) {
       \begin{minipage}{160bp}\kern0pt
         \bf Early arrival time 
       \end{minipage}
     };
  \node[ipe node, anchor=north west]
     at (384, 824) {
       \begin{minipage}{168bp}\kern0pt
         \bf Latest stabilization time 
       \end{minipage}
     };
  \node[ipe node, font=\Large]
     at (72.456, 792) {$\neg \val{a}{1} \wedge \neg \val{b}{1} $};
  \node[ipe node, font=\Large]
     at (228.176, 792) {$\val{a}{1} \wedge \val{b}{1} $};
  \node[ipe node, font=\Large]
     at (359.841, 792) {$\neg \val{a}{2} \wedge \neg \val{b}{2} $};
  \node[ipe node, font=\Large]
     at (504, 792) {$\val{a}{2} \wedge \val{b}{2} $};
  \draw[ipe pen fat]
    (48, 752)
     -- (80, 752)
     -- (80, 776)
     -- (136, 776);
  \draw[ipe pen fat]
    (48, 720)
     -- (104, 720)
     -- (104, 744)
     -- (136, 744);
  \draw[ipe pen fat]
    (136, 672)
     -- (104, 672)
     -- (104, 696)
     -- (48, 696);
  \draw[shift={(288, 720)}, xscale=-1, ipe pen fat]
    (0, 0)
     -- (56, 0)
     -- (56, 24)
     -- (88, 24);
  \draw[shift={(288, 752)}, xscale=-1, ipe pen fat]
    (0, 0)
     -- (32, 0)
     -- (32, 24)
     -- (88, 24);
  \draw[ipe pen fat]
    (200, 672)
     -- (232, 672)
     -- (232, 696)
     -- (288, 696);
  \draw[ipe pen fat]
    (344, 672)
     -- (376, 672)
     -- (376, 696)
     -- (432, 696);
  \draw[ipe pen fat]
    (432, 720)
     -- (400, 720)
     -- (400, 744)
     -- (344, 744);
  \draw[ipe pen fat]
    (432, 752)
     -- (376, 752)
     -- (376, 776)
     -- (344, 776);
  \draw[ipe pen fat]
    (576, 776)
     -- (544, 776)
     -- (544, 776)
     -- (488, 776);
  \draw[ipe pen fat]
    (576, 744)
     -- (544, 744)
     -- (520, 744)
     -- (488, 744);
  \draw[ipe pen fat]
    (488, 672)
     -- (520, 672)
     -- (544, 672)
     -- (576, 672);
  \filldraw[shift={(520, 776.038)}, xscale=0.7637, yscale=1.2504, fill=turquoise]
    (0, 0)
     -- (0, -96);
  \node[ipe node, font=\Large]
     at (24, 792) {$a)$ };
  \node[ipe node, font=\Large]
     at (176.195, 792) {$b)$ };
  \node[ipe node, font=\Large]
     at (320.133, 792) {$c)$ };
  \node[ipe node, font=\Large]
     at (464.149, 792) {$d)$ };
\end{tikzpicture}
}
\caption[x]{\protect Examples of the application of Eqs.~\ref{eq1}. The shaded areas denote the interval between the early and the latest stabilization time. In such time regions, the signal value is not determined, because we also account for multiple hazards. }
\label{waves}
\end{figure}

For instance, Eq.~\ref{eq1}a considers the case in which  the initial values of $ a$ and $ b$ are both 1. 
Since the NAND gate has output 1 whenever one of its inputs is 0, as soon as the
first of its inputs switches to 0, the output will leave its initial 0 value, as shown in Fig.~\ref{waves}b.
Value 0 is called a {\em controlling value} for the NAND gate, since as soon as one of its input takes the controlling value, the other inputs are irrelevant for the output.
Other gate types have a controlling value; we define this concept formally, on a generic logic gate indicated with the symbol $\odot$.
\begin{definition}[Controlling value]
\label{def:controlling_value}
The logic gate $\odot$ has a {\em controlling value} $c$ iff $ c \odot 1 = c \odot 0 = 1 \odot c = 0 \odot c$;
in such a case $\neg c$ is called {\em non-controlling value}.
\end{definition}
 Conversely, Eq.~\ref{eq1}b considers the situation in which both inputs have the controlling value 0. Therefore (Fig.~\ref{waves}a), $y$ will switch only when both inputs have switched. Eqs.~\ref{eq1}c and \ref{eq1}d handle the cases in which an input takes the controlling value and the other the non-controlling value: the output will switch when the input at the controlling value switches.    

Eq.~\ref{eq1}e-h are used to compute the latest stabilization time of the output signal $y$.
Eq.~\ref{eq1}e takes care of the case in which
$ a$ and $ b$ have 1 (non-controlling) as final values. 
The output stabilizes to 0 only when both the inputs are stable (Fig.~\ref{waves}d). Conversely, in Eq.~\ref{eq1}f, both inputs have a controlling final value 0. Therefore, $y$ will stabilize to its final value 1 as soon as one of such inputs has stabilized (Fig.~\ref{eq1}d). Eq.~\ref{eq1}g and h, instead, feature an input with a final controlling value and the other one with a non-controlling value, therefore, the output will stabilize to its final value only when the input at the controlling value has stabilized.    

In order to compute the times  $\tempo{y}{1}$ and $\tempo{y}{2}$,
we have to consider that the time interval $[\tempos{y}{1},\tempos{y}{2}]$ might be of zero or negative length.
For example, consider the case featuring a rising transition on $ a$ ($\val{a}{1}=0$ and $\val{a}{2}=1$)
and  a falling one on $ b$ ($\val{b}{1}=1$ and $\val{b}{2}=0$),
with the following timing: $\tempo{a}{1}=4$, $\tempo{a}{2}=5$ and $\tempo{b}{1}=1$, $\tempo{b}{2}=3$, also depicted in Fig.~\ref{no}. 
It is evident from Fig.~\ref{no} that there is no instant in which both inputs are  1 at the same time, which implies that 
the output $y$ remains stable at 1, meaning that
no hazard may be present at the output. Note that Eq.~\ref{eq1}d and Eq.~\ref{eq1}g produce  
$\tempos{y}{1}=\tempos{a}{1}=4$ and $\tempos{y}{2}=\tempos{b}{2}=3$, 
respectively. 
The apparent inconsistency $\tempos{y}{2} < \tempos{y}{1}$
signals that
there is no transition on ${ y}$, which can be thought as if ${ y}$ reaches its final value at time
$-\infty$, and that the initial value is maintained until time $+\infty$.
Steady states of a signal $s$ (hazard and transition free within the two considered input vectors) are encoded by $\tempo{s}{1}=T$, and
$\tempo{s}{2}=-1$. Here, $T$ 
 is a value larger than the maximum delay computed by \ac{STA}, effectively representing an infinite early arrival time   (the signal never switches from $\val{s}{1}$), while
$\tempo{s}{2}=-1$ signifies that the signal has already stabilized to  $\val{s}{2}$ when the first input vector is applied).
We detect this situation when the following variable $\eta_{ y}$ is true:
\begin{equation}
\eta_{ y} \leftrightarrow (\val{y}{1}=\val{y}{2}) \wedge (\tempos{y}{2}-\tempos{y}{1} \le 0) \ .  
\label{hazards}
\end{equation}
Such a variable holds true if and only if there is a potential hazard that does not appear because of circuit delays.

\tikzstyle{ipe stylesheet} = [
  ipe import,
  even odd rule,
  line join=round,
  line cap=butt,
  ipe pen normal/.style={line width=0.4},
  ipe pen heavier/.style={line width=0.8},
  ipe pen fat/.style={line width=1.2},
  ipe pen ultrafat/.style={line width=2},
  ipe pen normal,
  ipe mark normal/.style={ipe mark scale=3},
  ipe mark large/.style={ipe mark scale=5},
  ipe mark small/.style={ipe mark scale=2},
  ipe mark tiny/.style={ipe mark scale=1.1},
  ipe mark normal,
  /pgf/arrow keys/.cd,
  ipe arrow normal/.style={scale=7},
  ipe arrow large/.style={scale=10},
  ipe arrow small/.style={scale=5},
  ipe arrow tiny/.style={scale=3},
  ipe arrow normal,
  /tikz/.cd,
  ipe arrows, <->/.tip = ipe normal,
  ipe dash normal/.style={dash pattern=},
  ipe dash dotted/.style={dash pattern=on 1bp off 3bp},
  ipe dash dashed/.style={dash pattern=on 4bp off 4bp},
  ipe dash dash dotted/.style={dash pattern=on 4bp off 2bp on 1bp off 2bp},
  ipe dash dash dot dotted/.style={dash pattern=on 4bp off 2bp on 1bp off 2bp on 1bp off 2bp},
  ipe dash normal,
  ipe node/.append style={font=\normalsize},
  ipe stretch normal/.style={ipe node stretch=1},
  ipe stretch normal,
  ipe opacity 10/.style={opacity=0.1},
  ipe opacity 30/.style={opacity=0.3},
  ipe opacity 50/.style={opacity=0.5},
  ipe opacity 75/.style={opacity=0.75},
  ipe opacity opaque/.style={opacity=1},
  ipe opacity opaque,
]

\definecolor{red}{rgb}{1,0,0}
\definecolor{blue}{rgb}{0,0,1}
\definecolor{green}{rgb}{0,1,0}
\definecolor{yellow}{rgb}{1,1,0}
\definecolor{orange}{rgb}{1,0.647,0}
\definecolor{gold}{rgb}{1,0.843,0}
\definecolor{purple}{rgb}{0.627,0.125,0.941}
\definecolor{gray}{rgb}{0.745,0.745,0.745}
\definecolor{brown}{rgb}{0.647,0.165,0.165}
\definecolor{navy}{rgb}{0,0,0.502}
\definecolor{pink}{rgb}{1,0.753,0.796}
\definecolor{seagreen}{rgb}{0.18,0.545,0.341}
\definecolor{turquoise}{rgb}{0.251,0.878,0.816}
\definecolor{violet}{rgb}{0.933,0.51,0.933}
\definecolor{darkblue}{rgb}{0,0,0.545}
\definecolor{darkcyan}{rgb}{0,0.545,0.545}
\definecolor{darkgray}{rgb}{0.663,0.663,0.663}
\definecolor{darkgreen}{rgb}{0,0.392,0}
\definecolor{darkmagenta}{rgb}{0.545,0,0.545}
\definecolor{darkorange}{rgb}{1,0.549,0}
\definecolor{darkred}{rgb}{0.545,0,0}
\definecolor{lightblue}{rgb}{0.678,0.847,0.902}
\definecolor{lightcyan}{rgb}{0.878,1,1}
\definecolor{lightgray}{rgb}{0.827,0.827,0.827}
\definecolor{lightgreen}{rgb}{0.565,0.933,0.565}
\definecolor{lightyellow}{rgb}{1,1,0.878}
\definecolor{black}{rgb}{0,0,0}
\definecolor{white}{rgb}{1,1,1}

\begin{figure}
\begin{minipage}[b]{.33\textwidth}
\resizebox{\textwidth}{!}{\begin{tikzpicture}[ipe stylesheet]
\filldraw[turquoise, ipe pen fat]
    (124, 752) rectangle (124, 752);
  \draw[ipe pen fat]
    (168, 696)
     -- (136, 696)
     -- (104, 696)
     -- (48, 696);
  \node[ipe node, font=\Large]
     at (24, 760) {$$\bf a$ $};
  \node[ipe node, font=\Large]
     at (24, 728) {$$\bf b$$};
  \node[ipe node, font=\Large]
     at (24, 676) {$$\bf y $ $};
  \node[ipe node, font=\large]
     at (67.575, 639.977) {$\tempo{b}{1}  $};
  \node[ipe node]
     at (100, 640) {$\tempo{b}{2} $};
  \node[ipe node, font=\large]
     at (116, 640) {$\tempo{a}{1}  $};
  \filldraw[turquoise, ipe pen fat]
    (120, 776) rectangle (136, 752);
  \filldraw[shift={(72.623, 744)}, xscale=1.3074, yscale=0.9931, turquoise, ipe pen fat]
    (0, 0) rectangle (24, -24);
  \filldraw[fill=white]
    (103.9954, 744.1815)
     -- (103.9954, 656.1815);
  \filldraw[fill=white]
    (72.2164, 743.9622)
     -- (72.2164, 655.9622);
  \node[ipe node, font=\large]
     at (132, 640) {$\tempo{a}{2}  $};
  \draw[ipe pen fat]
    (48, 752)
     -- (120, 752)
     -- (120, 776)
     -- (168, 776);
  \draw[ipe pen fat]
    (168, 720)
     -- (104, 720)
     -- (104, 744)
     -- (48, 744);
  \filldraw[shift={(119.967, 754.339)}, xscale=-25.3298, yscale=1.116, fill=white]
    (0, 0)
     -- (0, -88);
  \filldraw[shift={(136.686, 775.59)}, xscale=2.5148, yscale=1.3549, fill=white]
    (0, 0)
     -- (0, -88);
\end{tikzpicture}
}
\caption[x]{\protect Example of timing waveforms featuring a potential hazard which, however, does not occur because of the relative timing of signals. }
\label{no}
\end{minipage}
\hfill
\begin{minipage}[b]{.62\textwidth}
\centering
\begin{tikzpicture}[xscale=.30, yscale=.12, every node/.style={scale=.38}]
  \draw [thick](1,11) -- (9,11);
  \draw [thick](1,14) -- (4,14);
  \draw [thick](1,19) -- (5,19);
  \draw [thick](1,22) -- (3,22);
  \draw [thick](1,25) -- (3,25);
  \draw [thick](1,28) -- (6,28);
  \draw [thick](1,29) -- (4,29);
  \draw [thick](1,34) -- (4,34);
  \draw [thick](1,35) -- (5,35);
  \draw [thick](1,4) -- (5,4);
  \draw [thick](1,40) -- (2,40);
  \draw [thick](1,43) -- (2,43);
  \draw [thick](1,44) -- (2,44);
  \draw [thick](1,49) -- (2,49);
  \draw [thick](1,7) -- (4,7);
  \draw [thick](1,8) -- (10,8);
  \draw [dashed](10,1) -- (10,49);
  \draw [thick](10,11) -- (15,11);
  \draw [dashed](11,1) -- (11,49);
  \draw [thick](11,8) -- (15,8);
  \draw [dashed](12,1) -- (12,49);
  \draw [thick](12,2) -- (15,2);
  \draw [dashed](13,1) -- (13,49);
  \draw [dashed](14,1) -- (14,49);
  \draw [dashed](2,1) -- (2,49);
  \draw [thick](2,38) -- (15,38);
  \draw [thick](2,38) rectangle (2,40);
  \draw [thick](2,41) -- (15,41);
  \draw [thick](2,41) rectangle (2,43);
  \draw [thick](2,44) rectangle (2,46);
  \draw [thick](2,46) -- (15,46);
  \draw [thick](2,49) -- (15,49);
  \draw [dashed](3,1) -- (3,49);
  \draw [thick](3,20) -- (15,20);
  \draw [thick](3,20) rectangle (3,22);
  \draw [thick](3,23) -- (15,23);
  \draw [thick](3,23) rectangle (3,25);
  \draw [dashed](4,1) -- (4,49);
  \draw [thick](4,14) rectangle (4,16);
  \draw [thick](4,16) -- (15,16);
  \draw [thick](4,29) rectangle (4,31);
  \draw [thick](4,31) -- (15,31);
  \draw [thick](4,32) -- (15,32);
  \draw [thick](4,32) rectangle (4,34);
  \draw [thick](4,5) -- (15,5);
  \draw [thick](4,5) rectangle (4,7);
  \draw [dashed](5,1) -- (5,49);
\draw [thick](5,35) rectangle (5,37);
  \draw [thick](5,37) -- (15,37);
  \draw [dashed](6,1) -- (6,49);
  \draw [thick](6,19) -- (15,19);
  \draw [thick](6,26) -- (15,26);
  \draw [thick](6,26) rectangle (6,28);
  \draw [dashed](7,1) -- (7,49);
  \draw [dashed](8,1) -- (8,49);
  \draw [dashed](9,1) -- (9,49);
  \fill [turquoise] (10,8) rectangle (11,10);
  \fill [turquoise] (5,17) rectangle (6,19);
  \fill [turquoise] (5,2) rectangle (12,4);
  \fill [turquoise] (9,11) rectangle (10,13);
  \draw (0,12) node[text centered, font=\huge] {\it k};
  \draw (0,15) node[text centered, font=\huge] {\it j};
  \draw (0,18) node[text centered, font=\huge] {\it i};
  \draw (0,21) node[text centered, font=\huge] {\it o};
  \draw (0,24) node[text centered, font=\huge] {\it p};
  \draw (0,27) node[text centered, font=\huge] {\it f};
  \draw (0,3) node[text centered, font=\huge] {\it n};
  \draw (0,30) node[text centered, font=\huge] {\it e};
  \draw (0,33) node[text centered, font=\huge] {\it h};
  \draw (0,36) node[text centered, font=\huge] {\it g};
  \draw (0,39) node[text centered, font=\huge] {\it d};
  \draw (0,42) node[text centered, font=\huge] {\it c};
  \draw (0,45) node[text centered, font=\huge] {\it b};
  \draw (0,48) node[text centered, font=\huge] {\it a};
  \draw (0,6) node[text centered, font=\huge] {\it m};
  \draw (0,9) node[text centered, font=\huge] {\it l};
  \draw (10,0) node[text centered, font=\huge] {8};
  \draw (11,0) node[text centered, font=\huge] {9};
  \draw (12,0) node[text centered, font=\huge] {10};
  \draw (13,0) node[text centered, font=\huge] {11};
  \draw (14,0) node[text centered, font=\huge] {12};
  \draw (2,0) node[text centered, font=\huge] {0};
  \draw (3,0) node[text centered, font=\huge] {1};
  \draw (4,0) node[text centered, font=\huge] {2};
  \draw (5,0) node[text centered, font=\huge] {3};
  \draw (6,0) node[text centered, font=\huge] {4};
  \draw (7,0) node[text centered, font=\huge] {5};
  \draw (8,0) node[text centered, font=\huge] {6};
  \draw (9,0) node[text centered, font=\huge] {7};
\end{tikzpicture}
\caption{\label{fig:waveform_primo_circuito} Waveforms computed in the example circuit of Fig.~\ref{staticta}, in the case of maximum delay. 
The colored areas represent timings in which the value of the signals is undetermined. 
Waveforms automatically drawn with ASPECT \citep{aspect}.}
\end{minipage}
\end{figure}

The gate propagation delay can be added in all cases that are different from a constant gate output. This is made by adding the following constraints:
\begin{subequations}
\begin{align}
(\tempos{y}{1}\neq T) \wedge \neg \eta_{ y} \ \rightarrow \ & (\tempo{y}{1}=\tempos{y}{1}+d) \label{delay_1}\\
(\tempos{y}{2}\neq -1) \wedge \neg \eta_{ y} \ \rightarrow \ & (\tempo{y}{2}=\tempos{y}{2}+d) \label{delay_2}\\
(\tempos{y}{1}= T) \vee \eta_{ y} \ \rightarrow \ & (\tempo{y}{1}=T) \label{delay_3}\\
(\tempos{y}{2}=-1) \vee \eta_{ y} \ \rightarrow \ & (\tempo{y}{2}=-1) \label{delay_4}
\end{align}
\label{delay}
\end{subequations}
In the example in Fig.~\ref{no}, $\eta_{ y}=1$ makes false conditions in Eqs.~\ref{delay_1}-\ref{delay_2} and true Eq.~\ref{delay_3}-\ref{delay_4},
obtaining that $\tempo{y}{1}=T$ (the initial value of $y$ is maintained until time $T$) and $\tempo{y}{2}=-1$ (the final value of $y$ is available since time $-1$), meaning that signal $y$ never changes.

A more complete example is shown in Fig.~\ref{fig:waveform_primo_circuito}, where the waveforms are computed using the described method for the circuit in Fig.~\ref{staticta}.

The constraints describing other kinds of gates (NOR, AND, OR and NOT) can be computed in a similar way. In non-monotonic gates such as XOR and XNOR, the early arrival time \tempos{{}}{1} of the output is equal to the minimum arrival time of the inputs and the latest stabilization time \tempos{{}}{2} is equal to the maximum stabilization time of the inputs.

 \section{Preliminaries}
\label{sec:preliminaries}

Answer Set Programming (ASP) is a form of declarative programming oriented towards difficult combinatorial search problems \citep{ApplicationsASP16,AnswerSetPlanning,IAJournal24}. It relies on the stable model semantics, also known as answer set semantics \citep{StableModelsSemantics}. An ASP program $\Pi$ consists of a finite set of rules, each of which is an implication of the form $H \mathtt{:-} B$, where $H$ is the head and $B$ is the body of the rule.

The head $H$ can be an atom, a choice atom, or an aggregate atom. An atom has the form $\mathtt{a(t_1,\dots,t_n)}$, where $\mathtt{t_1},\dots,\mathtt{t_n}$ are terms. A choice atom has the form $\mathtt{\{a(t_1,\dots,t_n)\}}$, and an aggregate atom can be of the form $A{\mathtt{\{t_1,\dots,t_m:c_1,\dots,c_m\}\circ n}}$, where $A$ can be \verb+#sum+, \verb+#min+, \verb+#max+, or \verb+#count+, $\circ$ is a relational operator and {\tt n} is an integer.

The body $B$ is a set of literals, which can be either positive ($\mathtt{a}$) or negative ($\mathtt{not~a}$), where {\tt a} can be an atom, an aggregate or a condition. Literals and rules containing no variables are called ground. The ground instantiation of a program $\Pi$, denoted as $gr(\Pi)$, consists of all ground instances of rules in $\Pi$.
A condition has the syntax $\mathtt{a(X) : c(X)}$, where $\mathtt{a(X)}$ is an atom and $\mathtt{c(X)}$ is a condition that must be satisfied. This allows for the instantiation of variables to collections of terms within a single rule. For example, the rule $\mathtt{q :- r(X) : p(X)}$ is expanded to a conjunction of $\mathtt{r(X)}$ for all $\mathtt{X}$ that satisfy $\mathtt{p(X)}$.

Rules with an empty body are called facts, while rules with an empty head are called \acp{IC}. The head of an IC is intended to be false.

An interpretation $I$ of a program $\Pi$ is a subset of the set of atoms occurring in $\Pi$. Atoms in $I$ are considered true, while all remaining atoms are false. The reduct $\Pi^I$ of a program $\Pi$ with respect to an interpretation $I$ is obtained by removing rules containing negative literals $\mathtt{not~a}$ where $\mathtt{a} \in I$, and removing all negative literals from the remaining rules. An interpretation $I$ is a stable model of $\Pi$ if it is a minimal model of $\Pi^I$.

ASP solvers typically work in two stages: grounding and solving. In the grounding stage, the program is converted into an equivalent ground program. The solving stage involves finding stable models (answer sets) of the ground program. \section{Problem Formalization in Answer Set Programming}
\label{sec:encodings}
\sloppy

\begin{figure}[htb]
\begin{minted}[fontsize=\scriptsize]{text}
signal(V) :- gate_in(V, _, _). |\label{lst:def_signal_begin}|
signal(V) :- gate_in(_, _, V). |\label{lst:def_signal_end}|
output_node(V) :- signal(V), not gate_in(_, _, V). |\label{lst:def_output_node}|
input_node(V) :- signal(V), not gate_in(V, _, _).  |\label{lst:def_input_node}|
boolean(0). boolean(1). |\label{lst:boolean}|
input_vec_no(1). input_vec_no(2). |\label{lst:input_vectors}|
1 = {v(V, InpVec, X) : boolean(X)} :- input_node(V), input_vec_no(InpVec). |\label{lst:value_choice}|
v(Y, InpVec, V) :- gate_in(Y, buff, A), v(A, InpVec, V). |\label{lst:buffer_value}\hfill|% Buffer gate
v(Y, InpVec, Vout) :- gate_in(Y, inv, A), v(A, InpVec, Vin), inv(Vin,Vout).  |\label{lst:inverter_value_begin}\hfill|% NOT gate (inverter)
inv(0, 1). inv(1, 0).                           |\label{lst:def_inv}| |\label{lst:inverter_value_end}|
|\neutralcode|((and;nand), 1). |\absorbing|((and;nand), 0). |\label{lst:neutral_absorbing_begin}|
|\neutralcode|((or;nor),   0). |\absorbing|((or;nor),   1). |\label{lst:neutral_absorbing_end}|
out_val(and, |\neutralcode|, 1). out_val(and, |\absorbing|, 0). |\label{lst:out_val_begin}|
out_val(or, |\neutralcode|, 0).  out_val(or, |\absorbing|, 1).
out_val(nand,Element,1-OutVal):- out_val(and,Element,OutVal).
out_val(nor, Element,1-OutVal):- out_val(or, Element,OutVal). |\label{lst:out_val_end}|
v(Y, InpVec, OutValue) :- |\label{lst:ctrl_input_present_begin}\hfill| % AND, OR, NAND, NOR with a |\absorbing| value in input 
    gate_in(Y, Gate, In), 
    |\absorbing|(Gate, |\Absorbing|),
    out_val(Gate,|\absorbing|,OutValue),
    v(In, InpVec, |\Absorbing|).           |\label{lst:ctrl_input_present_end}|
v(Y, InpVec, OutValue) :- |\label{lst:ctrl_input_absent_begin}\hfill| % AND, OR, NAND, NOR when all inputs are |\neutral| values
    gate_in(Y, Gate, _), input_vec_no(InpVec),
    |\neutralcode|(Gate, |\NeutralVar|),
    out_val(Gate, |\neutralcode|, OutValue),
    v(In, InpVec, |\NeutralVar|) : gate_in(Y, Gate, In). |\label{lst:ctrl_input_absent_end}|
v(Y, InpVec, OutValue) :- |\label{lst:xor_equal_begin}\hfill| % xor, xnor with same input                    
    gate_in(Y, Gate, A), gate_in(Y, Gate, B), A != B,
    out_on_equal(Gate, OutValue),
    v(A, InpVec, VA), v(B, InpVec, VB), VA == VB.       |\label{lst:xor_equal_end}|
v(Y, InpVec, InvOutValue) :- |\label{lst:xor_differ_begin}\hfill| % xor, xnor with different input           
    gate_in(Y, Gate, A), gate_in(Y, Gate, B), A != B,
    out_on_equal(Gate, OutValue), inv(OutValue, InvOutValue),
    v(A, InpVec, VA), v(B, InpVec, VB), VA != VB.       |\label{lst:xor_differ_end}|
out_on_equal(xor, 0). out_on_equal(xnor, 1).  |\label{lst:ineq}|
\end{minted}

\caption{\label{fig:encoding_signals}Encoding in ASP of signals and logic gates, common to all the encodings.}
\end{figure}

A combinational circuit 
is encoded in \ac{ASP} using two types of facts. 
Each logic gate is represented by a fact {\tt gate\_delay(Out, Gate, Delay)}, 
where {\tt Gate} is the type of the gate, which can be {\tt and}, {\tt nand}, {\tt or}, {\tt nor}, {\tt xor}, {\tt xnor}, {\tt inv} (inverter)\footnote{We did not use {\tt not} because it is a reserved word in ASP.} or {\tt buff} (buffer);
{\tt Out} is the output signal of the gate, and  is also used as a unique identifier of the gate, since 
a signal can be output only of one gate;
{\tt Delay} is the delay associated with the gate.

In order to represent uniformly gates with varying number of input signals,
a fact {\tt gate\_in(Out, Gate, Input)} specifies that the gate of type {\tt Gate}, whose output is {\tt Out}, receives {\tt Input} as one of the input signals. 
Each input is described through a separate fact, allowing gates to have an arbitrary number of inputs when applicable.
For instance, a 3-input NAND gate with inputs {\tt a}, {\tt b}, and {\tt c}, and output {\tt y} is represented by the following three facts: {\tt gate\_in(y, nand, a). gate\_in(y, nand, b). gate\_in(y, nand, c).}

To reason about signal behavior in the circuit, we define terms representing signals (Figure~\ref{fig:encoding_signals}). A signal is any node appearing as an input or output of at least one gate (lines \ref{lst:def_signal_begin}-\ref{lst:def_signal_end}).
From this, we automatically identify \acp{PI} and \acp{PO}. 
A signal is a PI if it is never found as a gate output (line~\ref{lst:def_input_node}), and as a PO if it is never used as a gate input (line~\ref{lst:def_output_node}). 
With these definitions, the inputs and outputs are automatically inferred from the gate-level structure, with no need for external annotations or manual specification; modern grounders can nevertheless convert these definitions into a set of facts, which are handled very efficiently.

We consider the behavior of a combinational circuit under a pair of input vectors as introduced in Section~\ref{sec:problem}.
We define a domain of Boolean values, in line \ref{lst:boolean}, and specify the two input vectors in line \ref{lst:input_vectors}.

As the objective is to find two input vectors that witness the required properties (e.g., longest delay),
we introduce in line \ref{lst:value_choice} a choice rule that selects exactly one Boolean value for each input vector
and each \ac{PI}.
The atom {\tt v(V, InpVec, X)} means that signal {\tt V} takes the Boolean value {\tt X} under input vector {\tt InpVec}.

For non-primary input signals, the logical value is determined by the functional behavior of the gates driving them;
again, each signal takes a boolean value for each input vector. 
In the case of unary gates, such as {\tt buff} (buffer) and {\tt inv} (inverter), the output value is  determined by the value of their single input signal (lines \ref{lst:buffer_value}-\ref{lst:inverter_value_end}).
These rules are applied for each input vector ({\tt InpVec} $\in {1, 2}$).

In order to define compactly the logic behavior of multi-input gates, avoiding long lists of cases,
we leverage on the concept of {\em \absorbing} and {\em \neutral}  values (Def.~\ref{def:controlling_value}).
Value 0 is \absorbing\ for AND and NAND, while 1 is their \neutral\ element (line~\ref{lst:neutral_absorbing_begin}).
Value 1 is  \absorbing\ for OR and NOR, while 0 is \neutral\ (line~\ref{lst:neutral_absorbing_end}).
Lines~\ref{lst:out_val_begin}-\ref{lst:out_val_end} provide the truth table of these four gate types relying only on the \absorbing\ and \neutral\ values.

The output signal values for gates belonging to types enjoying \absorbing\ values 
are computed in lines~\ref{lst:ctrl_input_present_begin}-\ref{lst:ctrl_input_absent_end}.
The first clause (lines \ref{lst:ctrl_input_present_begin}-\ref{lst:ctrl_input_present_end}) considers the case in which at least one input takes the \absorbing\ value;
the second (lines~\ref{lst:ctrl_input_absent_begin}-\ref{lst:ctrl_input_absent_end}) when all inputs take the \neutral\ value.
The conditional atom \mintinline{text}{v(In, InpVec, |\NeutralVar|) : gate_in(Y, Gate, In)} is expanded at grounding time into a conjunction of atoms \mintinline{text}{v(In, InpVec, |\NeutralVar|)}, one for each ground atom satisfying
\mintinline{text}{gate_in(Y, Gate, In)}; in the example of a NAND gate with inputs {\tt a}, {\tt b} and {\tt c}, 
it
is expanded to the conjunction \mintinline{text}{v(a, InpVec, |\NeutralVar|)}, \mintinline{text}{v(b, InpVec, |\NeutralVar|)}, \mintinline{text}{v(c, InpVec, |\NeutralVar|)}, so it is true only of all the three inputs take the \neutral\ value.

Finally, to simplify the exposition we show the code for modeling XOR and XNOR gates having exactly two inputs. 
The code can be extended also for multi-inputs, but is not necessary for
common gate libraries used in benchmarks, where XOR/XNOR gates are generally with 2 inputs.
The auxiliary predicate {\tt out\_on\_equal/2} (line~\ref{lst:ineq}) defines the output value of these gate types when the two inputs are equal.
A XOR gate produces 0 when both inputs are equal, whereas XNOR produces 1.
The logical behavior is encoded through two rules. The first  handles the case where the  inputs have the same value (lines \ref{lst:xor_equal_begin}-\ref{lst:xor_equal_end}), and the gate produces the output defined by {\tt out\_on\_equal/2}.
The second rule, lines \ref{lst:xor_differ_begin}-\ref{lst:xor_differ_end}, covers the case with different input values.

\begin{figure}
\begin{minted}[fontsize=\scriptsize,escapeinside=||]{text}
t(V, InpVec, 0) :- input_node(V), input_vec_no(InpVec). |\label{lst:input_time}|
eta(SignalY) :- fixed(SignalY), |\label{lst:def_eta_begin}\hfill|        % Definition of eta, see equation |(\ref{hazards})|
    ts(SignalY, 1, Ey), ts(SignalY, 2, Ly), Ly-Ey <= 0. |\label{lst:def_eta_end}|
fixed(SignalY) :- v(SignalY, 1, Vy), v(SignalY, 2, Vy).  |\label{lst:def_fixed}|
t(SignalY, 1, EyStar+Delay) :- EyStar != MaxTime, maxtime(MaxTime), |\label{lst:delay_1_2_begin}\hfill| % Eq (|\ref{delay_1}|)
    not eta(SignalY), ts(SignalY, 1, EyStar),  gate_delay(SignalY, _, Delay).
t(SignalY, 2, LyStar+Delay) :- LyStar != -1, |\hfill| % Eq (|\ref{delay_2}|)
    not eta(SignalY), ts(SignalY, 2, LyStar),  gate_delay(SignalY, _, Delay).                                  |\label{lst:delay_1_2_end}|
t(SignalY, 1, MaxTime) :- maxtime(MaxTime), ts(SignalY, 1, MaxTime).  |\label{lst:delay_3_begin}\hfill| % First  part of Eq (|\ref{delay_3}|)
t(SignalY, 1, MaxTime) :- maxtime(MaxTime), eta(SignalY). |\hfill| |\label{lst:delay_3_end}|% Second part of Eq (|\ref{delay_3}|)
t(SignalY, 2, -1) :- ts(SignalY, 2, -1).|\label{lst:delay_4_begin}\hfill|% First part of Eq (|\ref{delay_4}|)
t(SignalY, 2, -1) :- eta(SignalY). |\label{lst:delay_4_end}\hfill|% Second part of Eq (|\ref{delay_4}|)
static_timing_analysis(V, 0) :- input_node(V). |\label{lst:static_analysis_begin}|
static_timing_analysis(Out, MaxIn + Delay) :- gate_in(Out, Gate, In),    
	gate_delay(Out, Gate, Delay),  static_timing_analysis(In,MaxIn).|\label{lst:static_analysis_end}|
maxtime(MaxTime+1) :- #max{T: static_timing_analysis(V, T), output_node(V)} = MaxTime. |\label{lst:maxtime}|
max_output_delay(MaxOutDelay) :- MaxOutDelay = #max{L, OutNode : t(OutNode, 2, L), output_node(OutNode) }. |\label{lst:predicate_max_output_delay}|
#maximize{ MaxOutDelay: max_output_delay(MaxOutDelay) }. |\label{lst:optimization}|
\end{minted}
\caption{\label{fig:calcolo_tempi} Time computation and optimization}
\end{figure}

Figure~\ref{fig:calcolo_tempi} shows the code for computing the 
early ($e_{\bf v}$) and late ($l_{\bf v}$) arrival times of each signal according to the timing model
introduced in Section~\ref{sec:problem}.
Table~\ref{tab:symbols} summarizes how the variables in equations~\ref{eq1}-\ref{delay} are encoded in \ac{ASP}.
Since early and late arrival  are times associated to the first and second bit vector,
both are represented in \ac{ASP} through the same predicate {\tt t(Signal, InpVec, Time)}, where {\tt InpVec = 1} refers to the early arrival time and {\tt InpVec = 2} to the latest stabilization time.
For primary input nodes, these times are fixed to zero, as shown in line~\ref{lst:input_time}.

The auxiliary  variables $\tempos{{}}{1}$ and $\tempos{{}}{2}$ are represented by the predicate {\tt ts(Signal, InpVec, Time)};
{\tt ts} stands for {\em Time Star} and the parameters have the same meaning as in {\tt t}$/3$.

\begin{table}
\begin{center}
\begin{tabular}{ccl}
Symbol & ASP  & Description \\
\hline
$\tempo{v}{1}$ & {\tt Ev} in \mintinline{text}{t(v,1,Ev)} & Early Arrival Time of signal \segnale{v}\\
$\tempo{v}{2}$ & {\tt Lv} in \mintinline{text}{t(v,2,Lv)} & Late Arrival Time of signal \segnale{v}\\
$\tempos{v}{1}$ & {\tt Ev} in \mintinline{text}{ts(v,1,Ev)} & Auxiliary Early Arrival Time of signal \segnale{v}\\
$\tempos{v}{2}$ & {\tt Lv} in \mintinline{text}{ts(v,2,Lv)} & Auxiliary Late Arrival Time of signal \segnale{v}\\
$\eta_{\segnale{v}}$  & {\tt eta(v)} & Masked potential hazard in signal \segnale{v}\\
$T$ & {\tt maxtime(T)} & A time larger than the possible maximum delay \\
\hline
\end{tabular}
\end{center}
\caption{\label{tab:symbols} Symbols in the mathematical formulation (Sect.~\ref{sec:problem}) and in the ASP code.}
\end{table}

Lines~\ref{lst:def_eta_begin}-\ref{lst:def_eta_end} define $\eta$ as in Eq~\ref{hazards}.
Equations~\ref{delay_1}-\ref{delay_2} are implemented in lines~\ref{lst:delay_1_2_begin}-\ref{lst:delay_1_2_end}, where the arrival times are computed by adding the gate delay to the early (\tempos{y}{1}) and late (\tempos{y}{2}) auxiliary values. 
Eq.~\ref{delay_3} is implemented in lines~\ref{lst:delay_3_begin}-\ref{lst:delay_3_end} while Eq.~\ref{delay_4} corresponds to clauses~\ref{lst:delay_4_begin}-\ref{lst:delay_4_end}.

Equations~\ref{delay_1} and \ref{delay_3} require a value $T$ that is higher than the maximum delay;
while for correctness any value sufficiently large is valid, larger values can increase the solving time.
For this reason, we compute, in lines~\ref{lst:static_analysis_begin}-\ref{lst:static_analysis_end}, the upper bound with \ac{STA}. 
Probably, the most intuitive formulation would be to state that, for each gate $G$, the maximum possible output delay is the maximum of the inputs with added the delay of the gate:

\begin{minted}[linenos=false,fontsize=\small]{text}
static_timing_analysis(V, 0) :- input_node(V).
static_timing_analysis(Out, MaxIn + Delay) :-  gate_delay(Out, _, Delay),
    MaxIn = #max { Time : gate_in(Out, _, In), static_timing_analysis(In, Time) }.
\end{minted}

This version is correct, but contains a recursion through the \mintinline{text}{#max} aggregate;
this recursion hinders the optimization of the gringo grounder, that is unable to compute the upper bound at
grounding time, leaving it to be computed at solving time.
The formulation in lines~\ref{lst:static_analysis_begin}-\ref{lst:static_analysis_end}, instead, is stratified,
and the maximum time is computed directly by the gringo grounder. The size of the ground program is also significantly reduced.
In the formulation in lines~\ref{lst:static_analysis_begin}-\ref{lst:static_analysis_end}, it can be the case that for some gate the value of the maximum time is not unique; in order to obtain the global maximum it is enough to apply the \mintinline{text}{#max} aggregate, as shown in line~\ref{lst:maxtime}.

Finally, the objective function is to maximize the maximum delay (lines \ref{lst:predicate_max_output_delay}-\ref{lst:optimization}).

The rest of the code is devoted to computing the value of the auxiliary variables \tempos{y}{1} and \tempos{y}{2} for each of the
gates types, based on Equations~\ref{eq1}a-\ref{eq1}h; in the next sections, we first show an intuitive encoding, then an advanced encoding for this task.

\subsection{A first encoding}
\label{sec:encoding_base}

\begin{figure}[htb]
\begin{minted}[fontsize=\scriptsize,escapeinside=||]{text}
ts(SignalY, 1, MinE) :-|\label{lst:ts_base_early_noabsorbing}\hfill|% Early* Arrival time in AND, OR, NAND, NOR gates when there is
    gate_in(SignalY, Gate, _),|\hfill|% no |\absorbing| value in input, Eq (|\ref{eq1}a|)
    |\absorbing|(Gate, |\Absorbing|),
    #count{ I : gate_in(SignalY, Gate, I), v(I, 1, |\Absorbing|) } = 0,
    MinE = #min { T : gate_in(SignalY, Gate, I), t(I, 1, T) }.
ts(SignalY, 1, MaxE) :-|\label{lst:ts_base_early_absorbing}\hfill|% Early* Arrival time in AND, OR, NAND, NOR gates when there is
    gate_in(SignalY, Gate, _),|\hfill|% at least one |\absorbing| value in input, Eq (|\ref{eq1}b),(\ref{eq1}c),(\ref{eq1}d|)
    |\absorbing|(Gate, |\Absorbing|),
    #count{ I : gate_in(SignalY, Gate, I), v(I, 1, |\Absorbing|) } > 0,
    MaxE = #max { T : gate_in(SignalY, Gate, I), v(I, 1, |\Absorbing|), t(I, 1, T) }.
ts(SignalY, 2, MaxL) :-|\label{lst:ts_base_late_noabsorbing}\hfill|% Late* Arrival time in AND, OR, NAND, NOR gates when there is
    gate_in(SignalY, Gate, _),|\hfill|% no |\absorbing| value in input, Eq (|\ref{eq1}e|)
    |\absorbing|(Gate, |\Absorbing|),
    #count{ I : gate_in(SignalY, Gate, I), v(I, 2, |\Absorbing|) } = 0,
    MaxL = #max { T : gate_in(SignalY, Gate, I), t(I, 2, T) }.
ts(SignalY, 2, MinL) :-|\label{lst:ts_base_late_absorbing}\hfill|% Late* Arrival time in AND, OR, NAND, NOR gates when there is
    gate_in(SignalY, Gate, _),|\hfill|% at least one |\absorbing| value in input, Eq (|\ref{eq1}f),(\ref{eq1}g),(\ref{eq1}h|)
    |\absorbing|(Gate, |\Absorbing|),
    #count{ I : gate_in(SignalY, Gate, I), v(I, 2, |\Absorbing|) } > 0,
    MinL = #min { T : gate_in(SignalY, Gate, I), v(I, 2, |\Absorbing|), t(I, 2, T) }.
ts(SignalY, 1, EA) :- SignalA != SignalB, EB >= EA,|\label{lst:ts_base_early_xor}\hfill|% Early* Arrival time in XOR, XNOR gates
    t(SignalA, 1, EA),  gate_in(SignalY, (xor;xnor), SignalA),
    t(SignalB, 1, EB),  gate_in(SignalY, (xor;xnor), SignalB).
ts(SignalY, 2, LB) :- SignalA != SignalB, LB >= LA,|\label{lst:ts_base_late_xor}\hfill|% Late* Arrival time in XOR, XNOR gates
    t(SignalA, 2, LA),  gate_in(SignalY, (xor;xnor), SignalA),
    t(SignalB, 2, LB),  gate_in(SignalY, (xor;xnor), SignalB).
ts(SignalY,InpVec,TA):-|\label{lst:ts_base_time_unary}\hfill|% Early* and Late* Arrival time in unary gates
    t(SignalA,InpVec,TA),  gate_in(SignalY,(inv;buff),SignalA).
\end{minted}
\caption{\label{fig:basicEncoding} Basic Encoding}
\end{figure}

In the first encoding, Equations~\ref{eq1}a-\ref{eq1}h are generalized to the case of multiple inputs;
in order to reduce the number of cases we rely again on the concept of \absorbing\ value,
and the eight equations \ref{eq1}a-\ref{eq1}h  are rewritten as four clauses.
The clauses on lines~\ref{lst:ts_base_early_noabsorbing} and \ref{lst:ts_base_late_noabsorbing}
compute, respectively, the \tempos{y}{1} and \tempos{y}{2} 
arrival times in case there is no \absorbing\ value in input to the gate;
from Eq.~\ref{eq1}a, $\tempos{y}{1} = \min \{\tempo{i}{1}\}$ 
where $i$ is an input to the gate, while (Eq.~\ref{eq1}e) $\tempos{y}{2} = \max \{\tempo{i}{2}\}$.
Clauses~\ref{lst:ts_base_early_absorbing} and \ref{lst:ts_base_late_absorbing}, instead, consider the case in which at least one of the inputs is equal to the \absorbing\ value. In such a case, \tempos{y}{1}
is the maximum of the \tempo{i}{1}, where $i$ are the inputs taking the \absorbing\ value (Eq \ref{eq1}b , \ref{eq1}c, \ref{eq1}d), while \tempos{y}{2} is the minimum of the \tempo{i}{2} that are equal to the \absorbing\ value (Eq \ref{eq1}f , \ref{eq1}g, \ref{eq1}h).

Considering XOR and XNOR gates, clause~\ref{lst:ts_base_early_xor} computes the \tempos{y}{1}, while clause~\ref{lst:ts_base_late_xor} computes the \tempos{y}{2}.  Finally, clause~\ref{lst:ts_base_time_unary} takes care of unary gates (inverter and buffer).

\subsection{An Advanced encoding}
\label{sec:encoding_advanced}

The encoding proposed in Section~\ref{sec:encoding_base} implements correctly the equations \ref{eq1}a-\ref{eq1}h,
but it suffers from the fact that the number of clauses in the ground program can be very large.
Consider for example the clause in line~\ref{lst:ts_base_early_noabsorbing} (corresponding to Eq.~\ref{eq1}a);
to simplify the exposition let us consider a simplified version that holds only for a NAND gate with two inputs, only for the early arrival time (so we remove the parameter {\tt InpVec} from all predicates to simplify the exposition):
\begin{minted}[linenos=false,fontsize=\small]{text}
ts(Y, EA) :- v(A,1), v(B,1), A!=B,          % Both inputs are 1
    gate_in(Y, Gate, A), t(A, EA), % Compute EA = time of input A
    gate_in(Y, Gate, B), t(B, EB), % Compute EA = time of input B
    EA <= EB.
\end{minted}
Such a clause intuitively means that if both inputs are 1, then the early arrival time \tempos{\mathtt{Y}}{1} is the minimum of the two inputs \tempo{\mathtt{A}}{1} and \tempo{\mathtt{B}}{1}. If {\tt EA} and for {\tt EB} can possibly take $t$ values each, such a clause is grounded into $O(t^2)$ clauses. For a gate with $k$ inputs, the number of clauses worsens to $O(t^k)$.

However, one can observe that in order to compute the minimum, it is not strictly necessary to know the exact timing of {\em all} the input signals: it is necessary only to know the exact timing of the smallest one, plus it is needed to have a proof that all the other signals have a larger (or equal) time.
So, if we define a predicate {\tt tgeq(S, Time)} that is true when  signal {\tt S} has a time greater than or equal to\footnote{The acronym {\tt tgeq} stands for {\tt t}ime is {\tt g}reater than or {\tt eq}ual to.} {\tt Time},
we would be able to compute the minimum using only a linear number of clauses with respect to $t$, as follows:
\begin{minted}[linenos=false,fontsize=\small]{text}
ts(Y, EA) :- v(A,1), v(B,1), A!=B,
    gate_in(Y, Gate, A), t(A, EA), % Compute EA = time of input A
    tgeq(B,EA).                    % Ensure that EB >= EA
\end{minted}
Note that in this version variable {\tt EB} does not occur in the clause, as it is not necessary to know the exact value of \tempo{\mathtt{B}}{1}: it is only necessary to know that \tempo{\mathtt{B}}{1}'s value is higher than or equal to \tempo{\mathtt{A}}{1}, which is stated by the atom {\tt tgeq(B,EA)}.
Now, predicate {\tt tgeq(S,T)}, with the intuitive meaning that $\tempo{\mathtt{S}}{1} \geq {\mathtt{T}}$,  can be defined as:
\begin{minted}[linenos=false,fontsize=\small]{text}
tgeq(S, T) :- t(S, T).
tgeq(S, T-1) :- T>=0, tgeq(S, T).
\end{minted}
The first clause states that if $\tempo{\mathtt{S}}{1}  ={\mathtt{T}}$, then obviously $\tempo{\mathtt{S}}{1} \geq {\mathtt{T}}$, while the second says that if $\tempo{\mathtt{S}}{1}  \geq {\mathtt{T}}$,  then also $\tempo{\mathtt{S}}{1} \geq {\mathtt{T}}-1$. Note that predicate {\tt tgeq} is defined with two Horn clauses (which are handled very efficiently by modern ASP solver - it is worth to remember that Horn-SAT is a polynomially solvable fragment of SAT), and that as soon as the ASP solver infers that atom {\tt t(S,T0)} is true for some specific value {\tt T0}, the truth of {\tt t(S,T)} is inferred $\forall {\tt T} \leq {\tt T0}$ 
through the simple
unit propagation mechanism, in linear time.

\begin{figure}[htb]
\begin{minted}[fontsize=\scriptsize,escapeinside=||]{text}
tgeq(S, InpVec, V) :- t(S, InpVec, V).
tgeq(S, InpVec, V-1) :- V>=0, tgeq(S, InpVec, V).
ts(SignalY, 1, EA) :- t(SignalA, 1, EA), |\label{lst:advanced:joined_clauses1_start}|
    all_|\neutralcode|_except(SignalY,1,SignalA),
    tgeq(SignalB, 1, EA) : gate_in(SignalY, Gate, SignalB), SignalB != SignalA.|\label{lst:advanced:joined_clauses1_end}\hfill|% EA is the minimum
ts(SignalY, 1, EB) :- t(SignalB, 1, EB), |\label{lst:advanced:early_absorbing_start}|
    gate_in(SignalY, Gate, SignalB), v(SignalB, 1, V), |\absorbing|(Gate, V),
    not tgeq(SignalA, 1, EB + 1) : gate_in(SignalY, Gate, SignalA), v(SignalA, 1, V), SignalA != SignalB.|\label{lst:advanced:early_absorbing_end}|
ts(SignalY, 2, LA) :- t(SignalA, 2, LA),
    all_|\neutralcode|_except(SignalY,2,SignalA),
    not tgeq(SignalB, 2, LA + 1) : gate_in(SignalY, Gate, SignalB), SignalB != SignalA.
ts(SignalY, 2, LB) :- t(SignalB, 2, LB),
    gate_in(SignalY, Gate, SignalB), v(SignalB, 2, V), |\absorbing|(Gate, V),
    tgeq(SignalA, 2, LB) : gate_in(SignalY, Gate, SignalA), v(SignalA, 2, V), SignalA != SignalB.
ts(SignalY, 1, Min) :- t(SignalA, 1, Min),
    gate_in(SignalY, (xor;xnor), SignalA),
    gate_in(SignalY, (xor;xnor), SignalB),
    SignalA != SignalB, tgeq(SignalB, 1, Min).
ts(SignalY, 2, Max) :- t(SignalA, 2, Max),
    gate_in(SignalY, (xor;xnor), SignalA),
    gate_in(SignalY, (xor;xnor), SignalB),
    SignalA != SignalB, not tgeq(SignalB, 2, Max + 1).
all_|\neutralcode|_except(SignalY,BitVec,SignalA):-|\hfill|% all inputs to the gate having output SignalY take 
    input_vec_no(BitVec),|\hfill|% the |\neutral| value, except possibly SignalA
    gate_in(SignalY, Gate, SignalA), |\neutralcode|(Gate, V),
    v(SignalB, BitVec, V) : gate_in(SignalY, Gate, SignalB), SignalB != SignalA.
\end{minted}
\caption{\label{fig:advanced_encoding} Advanced encoding}
\end{figure}

Figure~\ref{fig:advanced_encoding} shows the computation of \tempos{{}}{1} and \tempos{{}}{2} in the advanced encoding; the complete code of the advanced encoding consists of the code in Figures~\ref{fig:encoding_signals}, \ref{fig:calcolo_tempi} and \ref{fig:advanced_encoding}.

The clause in lines~\ref{lst:advanced:early_absorbing_start}-\ref{lst:advanced:early_absorbing_end} considers the case in which there is at least one input taking the \absorbing\ value in the first bit vector: in such a case, the early arrival time $\tempos{Y}{1}$ of the output is the maximum among the early arrival times of the inputs at the \absorbing\ value.
This is implemented with a clause stating that the early arrival $\tempos{Y}{1}=\tempo{B}{1}$ for some signal $B$ taking the \absorbing\ value and with early arrival time \tempo{B}{1} that is higher than or equal to all other inputs having the \absorbing\ value.

Another improvement in this encoding can be found in the clause in lines~\ref{lst:advanced:joined_clauses1_start}-\ref{lst:advanced:joined_clauses1_end}. Consider equation~\ref{eq1}a: in the case of multiple inputs it could be interpreted as: if all the inputs take the \neutral\ value, then the time $\tempos{Y}{1}$ is the minimum of the early arrival time of the inputs. This could be implemented as
\begin{minted}[linenos=false,fontsize=\small]{text}
ts(Y, 1, EA) :-
    t(A, 1, EA), v(A, 1, |\NeutralVar|), |\neutralcode|(Gate, |\NeutralVar|),
    v(B, 1, |\NeutralVar|) :  gate_in(Y, Gate, B), B != A;
    tgeq(B, 1, EA) : gate_in(Y, Gate, B), B != A;
    gate_in(Y, Gate, A).
\end{minted}
i.e., if the input {\tt SignalA} has early arrival time lower than or equal to all other inputs, and all the inputs take the \neutral\ value in the first bit vector, then $\tempos{\mathtt{Y}}{1} = \tempo{\mathtt{A}}{1}$. On the other hand if there exists an input {\tt A} having early arrival time lower than or equal to all other inputs
but taking the \absorbing\ value (while all other inputs take the \neutral\ value), then again $\tempos{\mathtt{Y}}{1} = \tempo{\mathtt{A}}{1}$,  this time due to Eq.~\ref{eq1}d.
So, it is possible to strengthen the clause, removing the condition {\tt v(A, 1, \NeutralVar)},
as in lines~\ref{lst:advanced:joined_clauses1_start}-\ref{lst:advanced:joined_clauses1_end}.
With the strengthened clause, the ground program is smaller, and unit propagation can be applied more often, since the clause is shorter.
 \section{Experimental Results}
\label{sec:experiments}

In order to evaluate the running time of our method, we performed experiments on circuits from the ISCAS85~\citep{iscas85} and ITC99~\citep{css00} benchmark suites\footnote{In the ITC99 case, we considered the combinational fraction of these circuits.}.  
The output signal of a gate $G$ can be connected in input to a number of other gates; such a number is called {\em fan-out},
and it influences the delay associated to the gate $G$. In the experiments, we used for each gate a delay proportional to the fan-out of the gate;
of course the same methodology and encoding could be used with delays computed with other methodologies (e.g., hardware simulations).

We compared the basic encoding presented in Section~\ref{sec:encoding_base} and the advanced encoding in Section~\ref{sec:encoding_advanced}. 
Table~\ref{tab:results-table} reports the number of logic gates for each circuit, as well as the solving and grounding times, and the grounding size in MB.
The results were obtained using clingo 5.7.1 \citep{DBLP:journals/tplp/GebserKKS19}  on an AMD EPYC 9454 processor at 2.75GHz, with a maximum of 64GB of reserved memory. In the tests, we used clingo with only one core (without parallelism)
and imposed a timeout (T/O) of 10 hours (36,000s). As the objective is to compare the two ASP approaches, we do not report the instances in which both methods incurred into a timeout.

The second encoding provides speedups from 1.5$\times$ to 18$\times$. Instance C6288 is notoriously hard in the literature \citep{paperC6288}; it is a 16 bits $\times$ 16 bits multiplier that has an exponential number of paths, and many methods in the literature are not able to solve it in reasonable time.
While the first encoding is unable to solve it within 10 hours,
the advanced encoding can solve it in less than 4 hours.
A similar behavior is observed for instance B14, a subset of the Viper processor, but with a much larger grounding size that exceeds 12 GB for the basic encoding while remaining below 2 GB with the advanced encoding. This is likely due to the higher number of logic gates (8567 in B14 versus 1588 in C6288), which significantly increases the grounding complexity.

\begin{table}
    \centering
    \footnotesize
    \caption{Benchmark results for a set of logic circuits using the two \ac{ASP} encoding. For each circuit, the number of logic gates, solving and grounding times, as well as grounding sizes are reported. The final two columns show the maximum observed late output and the maximum delay from static timing analysis.}\label{tab:results-table}
    {\begin{tabular}{@{\extracolsep{\fill}}lr|rrr|rrr|rr}
        \hline
        \multicolumn{2}{c|}{} &
        \multicolumn{3}{c|}{Basic encoding} & 
        \multicolumn{3}{c|}{Advanced encoding} &
        \multicolumn{2}{c}{} \\
        name & 
        \makecell{\# logic\\gates} & 
        \makecell{Total\\time [s]} & 
        \makecell{grounding\\time [s]} & 
        \makecell{grounding\\size [MB]} & 
        \makecell{Total\\time [s]} & 
        \makecell{grounding\\time [s]} & 
        \makecell{grounding\\size [MB]} & 
        \makecell{max\\late\\output} & 
        \makecell{max\\STA\\delay} \\
        \hline
C432   &152 &1.21 &0.33 &51.4 &0.71 &0.29 &7.8 &71 &71\\ 
        C499   &320 &0.35 &0.08 &9.1 &0.23 &0.20 &3.7 &40 &40\\ 
        C880   &311 &7.98 &1.46 &25.4 &2.21 &1.11 &5.9 &72 &72\\ 
        C1355  &601 &28.95 &4.55 &94.9 &18.10 &2.71 &20.6 &76 &76\\ 
        C1908  &274 &297.32 &13.93 &17.6 &57.49 &8.00 &5.0 &106 &118\\ 
        C2670  &761 &56.07 &5.27 &41.0 &18.13 &9.13 &11.6 &108 &112\\ 
        C3540  &996 &1566.54 &27.82 &244.2 &86.50 &23.57 &42.0 &126 &136\\ 
        C5315  &1605 &331.32 &23.33 &145.1 &104.00 &37.49 &38.1 &134 &138\\ 
        C6288  &1588 &T/O &49.68 &1582.5 &13291.97 &240.52 &251.0 &382 &386\\ 
        C7552  &3510 &1272.73 &43.73 &611.3 &160.14 &88.65 &133.8 &126 &130\\ 
        \hline
        B11 &622    &21.78  &2.72   &54.8   &5.09   &1.71   &16.4   &92 &105\\
        B12 &963    &25.41  &6.99   &75.3   &7.92   &4.70   &25.9   &70 &90\\
        B13 &310    &0.43   &0.36   &2.5    &0.32   &0.28   &2.1    &35 &35\\
        B14 &8567   &T/O    &1183.08    &12729.1 &12006.65   &589.69 &1858.1 &256    &259\\
\hline

    \end{tabular}}
\end{table}

 \section{Related Work}
\label{sec:related}

The work most closely related to the current article is by
\cite{DBLP:conf/lpnmr/AndresSGSBS13}: they address a closely related problem in hardware design, namely the computation
of the longest path in a combinational circuit, in \ac{ASP}.
Their approach is based on choosing a gate in the circuit, which is the gate under test,
and restrict their attention only on the paths that pass through that gate. This can help avoiding, in some instances,
large parts of the circuit, resulting in a very quick search.
Afterwards, they find the longest path (varying the gate under test) such that all the gates in the path commute
when the input array switches from the first array to the second.
In order to increase further the speed of the computation, they adopt a multi-shot solving strategy \citep{DBLP:journals/tplp/GebserKKS19}.
Indeed, their solution is extremely quick in instances having a number of paths not exceedingly high.
Our approach is, in most instances, slower, but more precise, because it solves a harder problem, taking into consideration also the hazards.

\ac{ASP} was employed in a number of different applications in hardware design and verification.
\cite{DBLP:journals/ijrc/IshebabiMBGS09} address a problem of automated design for multiprocessor systems on FPGAs.
\cite{DBLP:conf/lpnmr/AndresGSHRG13} propose in \ac{ASP} an automated
system design approach for embedded computing systems. 

\cite{DBLP:journals/tplp/GavanelliNPB17} address a problem of designing an optical router on-chip  maximizing  parallelism while avoiding routing faults.
They provide and compare approaches based on \ac{ASP}, Constraint Logic Programming and Integer Linear Programming.

\cite{DBLP:journals/jpdc/BobdaYGIS18} tackle system-level synthesis for heterogeneous multi-processors on a chip using ASP. 
\ac{ASP} is also at the basis of a stream reasoning tool that was used for monitoring and scheduling in a semiconductor failure analysis lab \citep{DBLP:conf/lpnmr/MastriaPCPPS24}.

Digital hardware verification is a complex process spanning from logic synthesis to physical design. For synchronous systems, functional and timing verification can be decoupled; this work focuses on the latter, to ensure that no timing violations occur.

Initial efforts centered on \ac{STA} \citep{sta0,sta1,sta}, later evolving to address false path identification \cite{k1994}. Early methods employed multi-valued algebras and custom justification algorithms to handle hazards. More recent approaches leverage SAT solvers for signal justification \citep{monsoon} and target critical path identification in complex modules \citep{10412039}.

Alternative strategies include event-driven simulation to exploit higher clock frequencies while mitigating false paths \citep{8299562}, and machine learning techniques like association rule analysis to identify input vectors critical to reliability \citep{10457948}. However, these methods offer approximate rather than exact solutions. \section{Conclusions}
\label{sec:conclusions}

We proposed an ASP-based solution for computing the maximum delay in  combinational circuits, a key task in hardware design and verification. This work, a collaboration between logic programming and hardware design experts, adopts a fine-grained approach that considers hazards, unlike the mainstream method of finding the maximum sensitizable path. Experiments show our approach can solve instances considered very hard in the literature. 
Future work includes experimenting with \ac{ASP} modulo difference logic \citep{clingo-dl}.

\bibliographystyle{tlplike}
\bibliography{biblio,delay}

\end{document}